\documentclass{article}
\usepackage{main,amsmath,graphicx}
\usepackage{subfig}
\usepackage[inline]{enumitem}
\usepackage[belowskip=-15pt,aboveskip=0pt]{caption}

\title{
Extracting deep local features to detect manipulated images of human faces
}
\name{Michail Tarasiou, Stefanos Zafeiriou}
\address{Imperial College London}
\begin{document}
%
\maketitle
\begin{abstract}
Recent developments in computer vision and machine learning have made it possible to create realistic manipulated videos of human faces, raising the issue of ensuring adequate protection against the malevolent effects unlocked by such capabilities. In this paper we propose local image features that are shared across manipulated regions are the key element for the automatic detection of manipulated face images. We also design a lightweight architecture with the correct structural biases for extracting such features and derive a multitask training scheme that consistently outperforms image class supervision alone. The trained networks achieve state-of-the-art results in the {\it FaceForensics++} dataset using significantly reduced number of parameters and are shown to work well in detecting fully generated face images.
\end{abstract}

\section{Introduction}
\label{sec:intro}
After decades of exposure to successfully manipulated image content people no longer think of images as self-evidently portraying an accurate representation of reality. Up until recently, video content has been thought of as a reliable source of information, given that realistic tampering required a considerable amount of resources and in most cases was easily identifiable. However, this has changed with the rise of automated techniques enabling successful facial manipulations. While these techniques have the capacity to kickstart a revolution in computer graphics and digital content creation, if misapplied can definitely have applications of malevolent abuse and severe negative impacts on human rights standards (e.g. privacy, stigmatization, discrimination). In addition, previously successful image forensics methods do not generalize to the new techniques or to the artifacts produced from video compression which is commonly applied when a video is uploaded in social media platforms. Thus, it is critical to develop tools that help the automatic authentication of video content. In this paper we make the following contributions towards that goal:
\begin{itemize}
    \item we propose that manipulation artifacts are local in nature, i.e. there is no high level semantic disparity between real and manipulated images, and are shared across the manipulated regions of an image. Following these principles we design a Convolutional Neural Network outperforming all networks of similar size or smaller in the {\it FaceForensics++} ({\it FF++}) dataset
    \item we show that adding auxiliary dense classification tasks improves model performance significantly over the base for all examined cases and derive a training scheme with similar gains that does not require having real forgery masks. This leads to state-of-the-art results in the {\it FF++} dataset while at the same time the number of model parameters is reduced twenty-fold
\end{itemize}.

\begin{figure*}[t]
\centering
\centerline{\includegraphics[trim={4cm 4cm 4cm 4cm}, width=16cm]{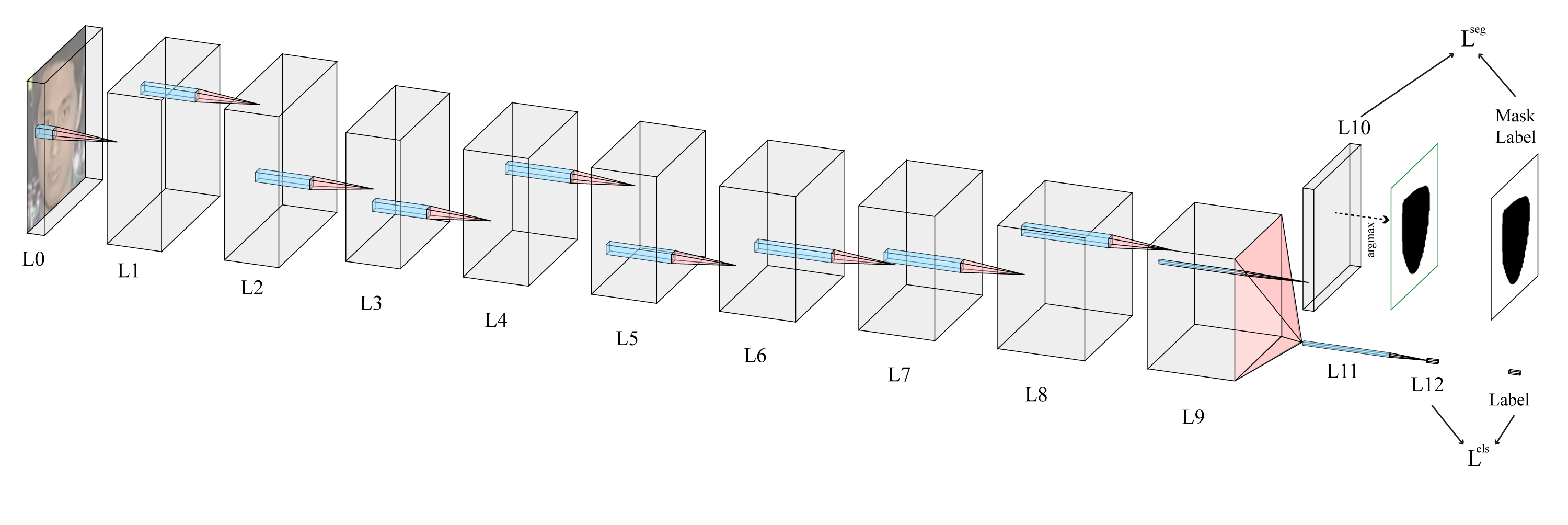}}
\caption{Network Architecture with a single segmentation loss $(L^{seg})$ component}
\label{net_arch}
\end{figure*}

\section{Related Work}
\label{sec:format}
Facial image manipulation techniques can broadly be categorized by two factors: \begin{enumerate*}
    \item the type of manipulation, namely identity or facial expression,
    \item whether the system uses computer vision based techniques or machine learning
\end{enumerate*}. {\it FaceSwap} \cite{fs} is a computer vision based technique for identity manipulation, while {\it Deepfakes} \cite{df} use deep autoencoders to achieve the same task. {\it Face2Face} \cite{f2f} is a computer vision system for live facial reenactment. Generative Adversarial Networks (GANs) \cite{gan} were introduced to model and produce new samples representative of a data distribution. Applied on human faces GANs can create highly realistic results in high resolution \cite{pgan, stylegan, tpdne} and perform image-to-image translations  for  multiple  domains showing impressive results in tasks such as altering the hair color, gender, age, skin tone and emotion of human faces.\\
Similarly manipulation detection techniques can be split between those employing handcrafted and learnt features. Image forensics and steganalysis focus on detecting a set of predefined manipulation artifacts that result from key image manipulation operations e.g. copy-move, splicing. They generally work great for a single type of forgery but fail in unseen cases and have been shown to deteriorate significantly in performance in the presence of video compression \cite{facefor++}.
More recent literature includes many works using deep learning methods. End-to-end trained CNNs have been shown to generally outperform other approaches. Results from training and testing various models provided in literature \cite{rahmouni, bayar, meso} is presented in \cite{facefor++}. All methods are shown to achieve high accuracy for uncompressed images but performance drops greatly when compression is introduced. In \cite{cnnrnn} the authors use temporal models on CNN features. For high compression a DenseNet architecture achieves state of the art precision on the binary classification task. The intuition that the final image rescaling and alignment steps common in various {\it deepfakes} pipelines should introduce face warping artifacts is further explored in \cite{warpdet} who propose a self-supervised technique bypassing the need for expensive positive examples. \cite{iio} take advantage of a bias of facial datasets not to include faces with their eyes closed to detect unrealistic blinking patterns in fake videos, while \cite{headpose} detect had-pose inconsistencies. \cite{world_leaders} build person specific models of their speaking mannerisms. Their models work well even in high compression but are person specific and thus can only be applied to specific individuals.

\begin{figure}
    \centering
    \subfloat[]{{\includegraphics[width=4cm]{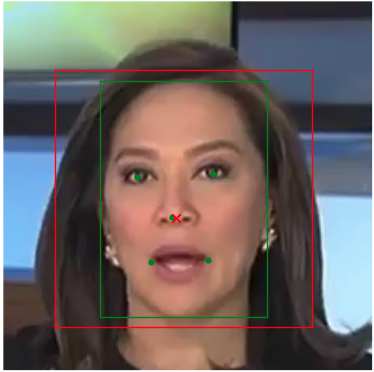} }}
    \subfloat[]{{\includegraphics[width=4cm]{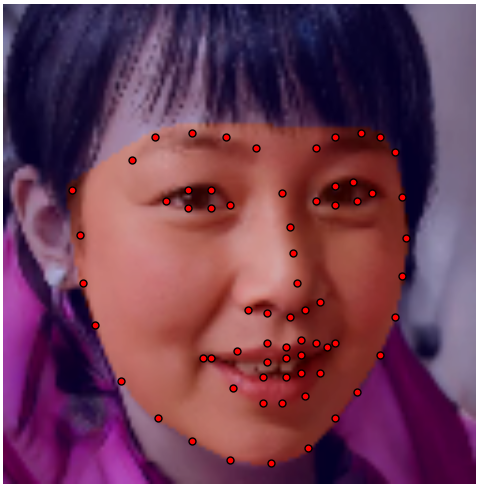} }}
    \caption{Data Preprocessing. (a) Face crop. Output of face detector in green, extracted face roi in red, (b) {\it Convex Hull Mask}. Face landmarks and convex-hull region in red}%
    \label{fig:data_preprocess}%
\end{figure}

\section{Proposed Approach}
\label{sec:pagestyle}
\subsection{Data Preprocessing}
\label{sec:data_preproc}
Incorporating prior knowledge into the data preprocessing pipeline has been shown to improve the performance of trained models \cite{facefor++}. For all the experiments that follow data preprocessing involves training on centered face crops rather than full images as shown in Fig.\ref{fig:data_preprocess}(a). When forgery masks are not available, we extract masks artificially as: \begin{enumerate*}
    \item {\it Zeros Mask} (ZM) when all image pixels are real, i.e. real images, 
    \item {\it Ones Mask} (OM) when all image pixels are generated, i.e. GAN images, 
    \item {\it Convex Hull Mask} (CVM) in which case we assign a positive class only to pixels inside the convex hull defined by facial landmarks (Fig.\ref{fig:data_preprocess}(b))
\end{enumerate*}. 

\subsection{Model Architecture}
In designing the model architecture presented below we follow the intuition that useful features for the classification problem at hand are non-global, non-semantic in nature. Visual examination of manipulated images does not offer sufficient clues. In a human study conducted in \cite{facefor++} humans were shown to be well inferior of most of the models tested, suggesting that successfully manipulated images cannot be identified based on appearance. Instead we assume that useful detection signal has two distinct characteristics: it can be detected in local image patches and it is consistent across the manipulated regions of an image. We design an architecture capable of capturing those features by respectively making the following choices: \begin{enumerate*}
    \item we restrict the final layer receptive field to 33 pixels or ($\approx1/16$ of input area), ensuring that no global features are used by the network,
    \item we use {\it Global Average Pooling} (GAP) \cite{detcg} across spatial locations for the final layer activations, forcing the model to extract features that are shared across image patches rather than patch specific features 
\end{enumerate*}.  
The proposed network architecture is illustrated in Fig.\ref{net_arch}. It consists of a feature extractor with 8 convolutional and one {\it MaxPool} layers (layers 1-9) and two or more separate classifiers (layers 10, 12) followed by {\it SoftMax} to derive class probabilities. All convolution steps are followed by {\it ReLU} activation \cite{relu} and Batch Normalization \cite{bn}. Convolutional filter strides are set to 1 and {\it MaxPool} stride is 2 for downsampling. No padding is used throughout the network.  

\subsection{Loss Function}
To further enforce the contribution of local features we use a joint classification-segmentation objective function. For the segmentation tasks we use real segmentation labels where available or construct approximations as discussed in section \ref{sec:data_preproc}. To extract segmentation labels for training we choose the indices corresponding to pixels at the center of each final layer activation's receptive field. For a mask label $y_{ij}$ and predicted class probabilities $p_{ij}$ at location $(i, j)$ the segmentation loss is defined as the average {\it Cross-Entropy Loss} $H(y_{ij}, p_{ij})$ over all locations $(i, j) \in \{1,...,N\}$:

$$L^{seg} = \frac{1}{N^2} \sum_{i}\sum_{j}H(y_{ij}, p_{ij})$$

At the same time we apply GAP and a linear layer followed by {\it SoftMax} to derive class probabilities for the image $p$. Similarly, these are used to define a {\it Cross-Entropy Loss} for labels $y$ and predicted probabilities $p$:
$$L^{cls} = H(y, p)$$

The joint loss function for the model is defined  for $k$ segmentation objectives:
$$L = \lambda^{cls} L^{cls} + \sum_{k}\lambda^{seg}_k L^{seg}_k$$
using hyperparameters:
$$\lambda^{cls}, \lambda^{seg}_k : 0 \leq \lambda^{cls}, \lambda^{seg}_k \leq 1.0, \lambda^{cls} + \sum_k \lambda^{seg}_k = 1.0$$

\section{Experiments}
\label{sec:typestyle}

\subsection{Datasets}
In the ablation study presented in section \ref{sec:forgloss_exp} we use the {\it FF++} dataset \cite{facefor++} which extends {\it FaceForensics} \cite{facefor} to include samples generated from {\it Deepfakes} \cite{df} and {\it FaceSwap} \cite{fs} containing 1000 videos per manipulation method.
Additionally, in section \ref{sec:ffd_exp} we use the {\it DFDD} \cite{dfd}, {\it DFDC} \cite{dfdc} and {\it Celeb-DF} \cite{celebdf} large scale deepfakes datasets.
For the GAN images detection experiments, pretrained networks based on \cite{pgan} and \cite{stylegan} were used to generate images of faces. From each technique 70k faces were selected such that a face could be found using a reasonably high detection threshold in a pretrained face detector \cite{retina_face} and these were split into 56k, 7k, 7k train, validation and test sets respectively. For real images the {\it Flickr-Faces-HQ} (FFHQ) dataset \cite{stylegan} of human faces was used. In order to examine the effect of compression on the detection performance all images were compressed using H.264 coding at levels similar to the ones used in the {\it FF++} dataset.

\begin{table}[ht]
\caption{Binary Classification Accuracy - (TOP) Manipulation Method-specific Training, (BOTTOM) Training with All Manipulation Methods}
\label{cls_acc}
\begin{center}
\begin{tabular}{|c|c|c|c|c|c|c|c|}
\hline
& \multicolumn{3}{c|}{C23 Compression} & \multicolumn{3}{c|}{C40 Compression} \\ 
\hline
$\lambda^{seg}$ & DF & F2F &  FS & DF & F2F &  FS \\
\hline
0.0 & 96.80 & 97.72 & 97.57 & 91.60 & 84.47 & 89.72\\
0.2 & \textbf{97.90} & 98.37 & 97.90 & 91.68 & 84.88 & 89.69 \\
0.3 & 97.86 & \underline{\textbf{98.58}} & \underline{\textbf{98.32}} & \textbf{92.40} & 86.32 & 89.77 \\
0.4 & 97.86 & 98.44 & 98.10 & 91.95 & 86.20 & 90.15 \\
0.5 & 97.75 & 98.43 & 98.24 & 91.80 & 86.40 & 90.23 \\
0.6 & 97.55 & 98.45 & 98.09 & 91.83 & \textbf{87.11} & 90.56 \\
0.7 & 97.51 & 98.38 & 98.17 & 91.71 & 86.81 & \textbf{91.26} \\
\cite{bayar} & 90.18 & 94.93 & 93.14 & 80.95 & 77.30 & 76.83 \\
\cite{meso} & 95.26 & 95.84 & 93.43 & 89.52 & 84.44 & 83.56 \\
\cite{facefor++} & \underline{98.85} & 98.36 & 98.23 & 94.28 & 91.56 & 93.70 \\
\cite{cnnrnn} & - & - & - & \underline{96.70} & \underline{93.21} & \underline{95.80} \\
\hline
0.0 & 94.54 & 94.73 & 94.16 & 85.92 & 82.89 & 83.49 \\
0.2 & 95.64 & 96.28 & 95.41 & 86.44 & 83.05 & 84.27 \\
0.3 & 95.77 & 96.23 & 95.97 & 85.96 & 83.96 & 84.82 \\
0.4 & 95.81 & 96.79 & 95.51 & \textbf{87.06} & 83.52 & 84.55 \\
0.5 & 96.14 & 96.54 & 95.96 & 86.23 & 83.74 & 84.19 \\
0.6 & \textbf{96.78} & 96.91 & \textbf{95.97} & 85.88 & 84.70 & \textbf{85.17} \\
0.7 & 96.65 & \textbf{97.13} & 95.92 & 86.83 & \textbf{84.98} & 84.77 \\
\cite{bayar} & 90.25 & 93.96 & 87.74 & 86.93 & 83.66 & 74.28 \\
\cite{meso} & 89.55 & 88.60 & 81.24 & 80.43 & 69.06 & 59.16 \\
\cite{facefor++} & \underline{97.49} & \underline{97.69} & \underline{96.79} & \underline{93.36} & \underline{88.09} & \underline{87.42}  \\
\hline
\end{tabular}
\end{center}
\end{table}

\subsection{Effect of Forgery Localization Loss}
\label{sec:forgloss_exp}
\begin{figure}[htb]
\begin{minipage}[b]{1.0\linewidth}
  \centering
  \centerline{\includegraphics[width=5cm]{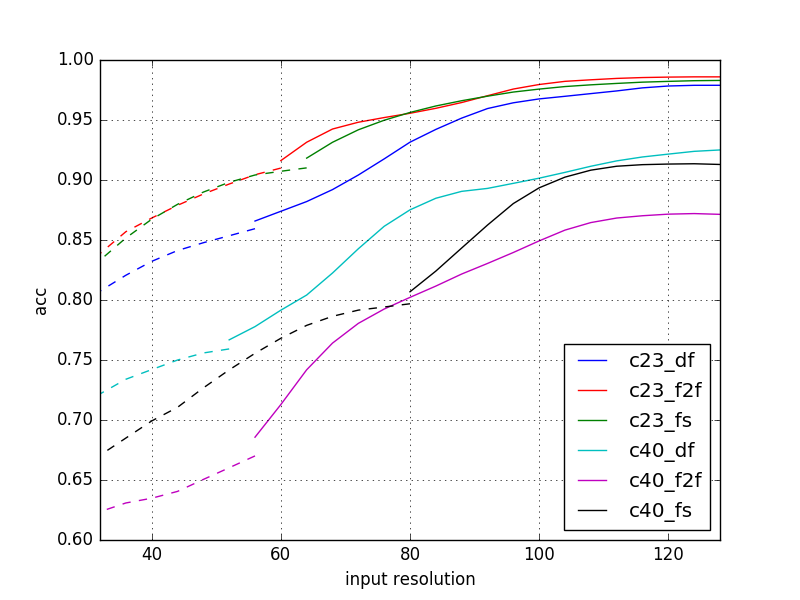}}
  \centerline{(a) Inference Classification accuracy with variable input size}\medskip
\end{minipage}
\begin{minipage}[b]{.48\linewidth}
  \centering
  \centerline{\includegraphics[trim={0cm 0cm 0cm 5cm}, width=4.5cm]{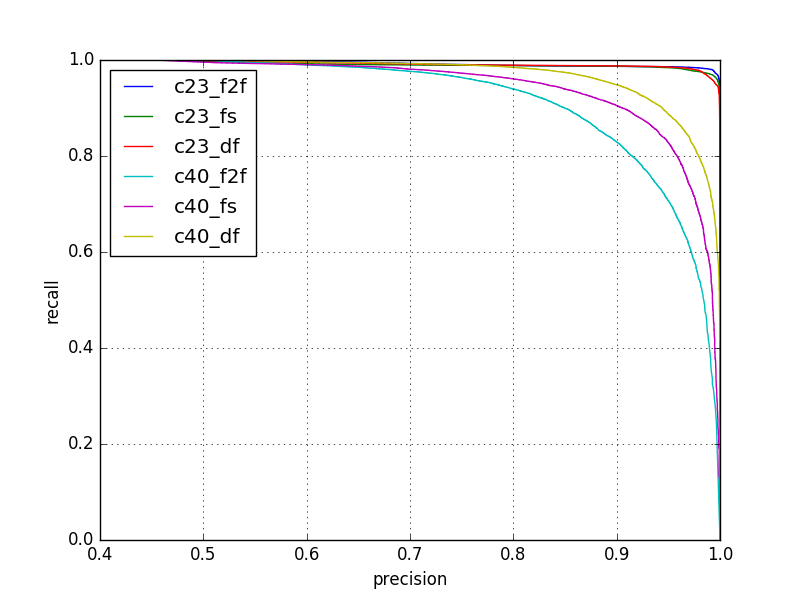}}
  \centerline{(b) Precision-Recall}\medskip
\end{minipage}
\hfill
\begin{minipage}[b]{0.48\linewidth}
  \centering
  \centerline{\includegraphics[width=4.5cm]{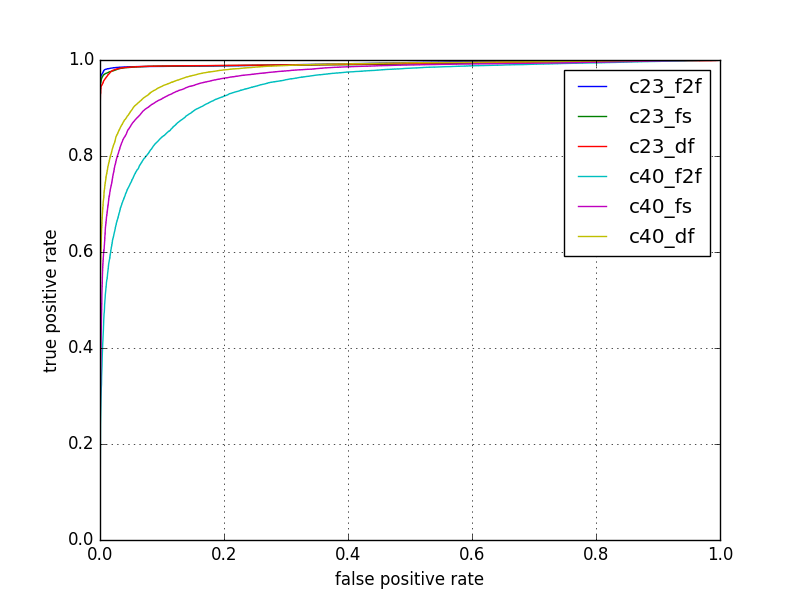}}
  \centerline{(c) ROC}\medskip
\end{minipage}
\caption{Manipulation detection - Test Metrics}
\label{fig:add_perf_metrics}
\end{figure}

To assess the contribution of forgery localization on classification performance we formulate our loss function using a single segmentation objective: to predict manipulated pixels. We perform an ablation study varying $\lambda_{seg}$ while using $\lambda_{seg}=0.0$ as the base case. Test set classification accuracies on {\it FF++} can be seen in Table \ref{cls_acc}. Overall there is a noticeable performance boost with up to $+2.6\%$ improvement over the base case which is always the worst performer. We note that all models work well for {\it c23} compression and achieve state of the art results (underlined numbers) using a significantly smaller network ($\sim 1M$ compared to $\sim 22M$ trainable parameters). For higher compression we note a significant performance deterioration for all models. We also note a performance drop in every case when the training set includes samples from all manipulation methods as opposed to method-specific training. Because our model does not employ fully connected layers it can be trained at a specified resolution and use variable size input images during inference. In practice this is performed by cropping the $128\times128$ original input. Reducing image size leads to a graceful performance degradation which introduces a trade-off between inference speed and performance (Fig.\ref{fig:add_perf_metrics}(a)). Dotted and solid lines correspond to using the mean probability of pixel forgery from the segmentation prediction - which works better in the small resolution regime - and the output of the image classifier. Additional performance metrics are presented in Fig.\ref{fig:add_perf_metrics}(b-c).

Per class classification accuracies for detecting generated images can be seen in Table \ref{gan_acc}. We notice that all networks perform better in this task and that they perform worse for the more realistic {\it StyleGan} images. Here, including segmentation loss using OMs was not found to improve results, however, using CVMs we see a noticeable jump in performance in every case. 
\begin{table}[t]
\caption{Binary Classification Accuracy - (TOP) Manipulation Method-specific Training, (BOTTOM) Training with All Manipulation Methods. (LEFT) OM, (RIGHT) CVM}
\label{gan_acc}
\begin{center}
\setlength\tabcolsep{2.7pt}
\begin{tabular}{|c|c|c|c|c|c|c|c|c|}
\hline
& \multicolumn{4}{c|}{C23 Compression} & \multicolumn{4}{c|}{C40 Compression} \\ 
\hline
$\lambda^{seg}$ & \multicolumn{2}{c|}{PGAN} & \multicolumn{2}{c|}{STYLEGAN} &  \multicolumn{2}{c|}{PGAN} & \multicolumn{2}{c|}{STYLEGAN} \\
\hline
0.0 & 99.96 & 99.96 & 99.55 & 99.55 & 99.76 & 99.76 & 97.66 & 97.66\\
0.2 & 99.97 & 99.97 & 99.55 & 99.77 & 99.77 & 99.87 & 97.59 & 98.31\\
0.3 & 99.98 & \textbf{99.99} & 99.59 & 99.78 & 99.81 & 99.93 & 97.60 & \textbf{98.68} \\
0.4 & 99.96 & \textbf{99.99} & 99.55 & 99.87 & 99.73 & 99.93 & 97.32 & 98.28\\
0.5 & 99.94 & \textbf{99.99} & 99.52 & 99.87 & 99.78 & 99.91 & 97.28 & 98.51\\
0.6 & 99.97 & \textbf{99.99} & 99.56 & \textbf{99.90} & 99.74 & \textbf{99.95} & 97.43 & 98.56 \\
0.7 & 99.94 & 99.98 & 99.50 & 99.85 & 99.74 & \textbf{99.95} & 97.27 & 98.52 \\
\hline
0.0 & 99.66 & 99.66 & 99.23 & 99.23 & 97.31 & 97.31 & 95.04 & 95.04\\
0.2 & 99.62 & 99.78 & 99.32 & 99.54 & 97.21 & 97.79 & 94.78 & 96.42 \\
0.3 & 99.63 & 99.88 & 99.31 & 99.60 & 97.33 & \textbf{98.59} & 94.89 & \textbf{97.46} \\
0.4 & 99.65 & 99.86 & 99.32 & 99.68 & 97.02 & 98.21 & 94.94 & 96.98 \\
0.5 & 99.53 & 99.87 & 99.33 & 99.70 & 97.17 & 98.27 & 94.75 & 96.93 \\
0.6 & 99.61 & 99.86 & 99.20 & \textbf{99.80} & 96.99 & 98.41 & 94.84 & 97.29 \\
0.7 & 99.65 & \textbf{99.89} & 99.24 & 99.79 & 97.03 & 98.53 & 94.89 & 97.28 \\
\hline
\end{tabular}
\end{center}
\end{table}

\begin{table}[h!]
\caption{Binary Classification Accuracy - Different Segmentation Objectives}
\label{segm_obj}
\begin{center}
\setlength\tabcolsep{3.5pt}
\begin{tabular}{|c|c|c|c|c|c|c|c|}
\hline
Case & \multicolumn{2}{|c|}{Mask} & \multicolumn{4}{c|}{Dataset}  \\ 
\hline
$\#$ & Face & Fake & FF-DF & DFDD & DFDC \footnotemark & Celeb-DF\\
\hline
1 & None & None & 96.80 & 94.41 & 85.76 & 91.05\\
2 & None & Real & 97.90 & \textbf{97.14} & - & 92.57\\
3 & None & CVM  & 97.67 & 96.82 & 88.20 & 92.29 \\
4 & CVM & None  & 97.20 & 95.30 & 83.26 & 90.79 \\
5 & CVM & Real  & 97.74 & 96.92 & - & 91.42 \\
6 & CVM & CVM   & \textbf{98.03} & 97.03 & \textbf{88.76} & \textbf{92.62} \\
7 & Real & Real & 97.97 & 96.16 & - & 91.50 \\
\hline
\end{tabular}
\end{center}
\end{table}

\footnotetext{No real forgery masks are provided for {\it DFDC}}

\subsection{Effect of Segmentation Objective}
\label{sec:ffd_exp}
In section \ref{sec:forgloss_exp} we showed that networks trained to detect fully generated images benefit from learning to separate foreground face pixels from the background even though all pixels in these images are generated. To assess whether these benefits can be attributed to learning to detect manipulated pixels or to separate face pixels from background and how well CVMs can replace real forgery masks in the proposed training scheme we compare results from the following experiments using the {\it FF++}, {\it DFDD}, {\it DFDC} and {\it Celeb-DF} datasets: \begin{enumerate*}
    \item base case from section \ref{sec:forgloss_exp},
    \item best model from section \ref{sec:forgloss_exp},
    \item use CVMs instead of real forgery masks to predict manipulated pixels,
    \item use CVMs to predict face region instead of manipulated pixels,
    \item use CVMs to predict face region and real masks to predict manipulated pixels $(k=2)$.
    \item use CVMs to predict face region and manipulated pixels $(k=2)$,
    \item use real masks to predict face region and manipulated pixels $(k=2)$
\end{enumerate*}.
Results are presented in Table \ref{segm_obj}. We note that learning to detect manipulated pixels appears to be the more important task, however, including both segmentation objectives is shown to work better overall. In every case including a segmentation component in the loss function boosts classification performance, except for case 4 where for the more challenging {\it DFDC} and {\it Celeb-DF} datasets it underperforms the base case. Surprisingly, using CVMs for segmenting both face and manipulated pixels appears to work better than using real segmentation masks for both tasks. This could be explained by the fact that CVMs cover a larger area than real masks and thus cover the information rich boundary between manipulated and non-manipulated pixels. If a single objective were to be selected that would be case 2 which was the training scheme used in experiments presented in section \ref{sec:forgloss_exp}. 
\section{Conclusion}
\label{sec:majhead}
In this paper we examined local features in the task of detecting manipulated images of human faces and proposed an convolutional architecture and training scheme that target finding such features. Adding dense localization components in the objective function was shown to boost classification performance. Using that approach we achieved a strong improvement over our baselines and state-of-the-art results in the {\it FF++} dataset. Finally, we proposed a methodology for automatically generating all required training labels eliminating the need for real-fake video pairs. 


\bibliographystyle{IEEE}
\raggedright
\bibliography{main}

\end{document}